\documentclass{article}
\PassOptionsToPackage{numbers,square,sort&compress}{natbib}
\usepackage[dblblindworkshop,final]{neurips_2026}
\workshoptitle{The 6th Workshop on Mathematical Reasoning and AI}
\usepackage[utf8]{inputenc}
\usepackage[T1]{fontenc}
\usepackage{hyperref}
\usepackage{url}
\usepackage{booktabs}
\usepackage{amsmath,amssymb}
\usepackage{graphicx}
\usepackage{microtype}
\usepackage{xcolor}

\hypersetup{hidelinks,pdfauthor={Guilin Zhang, Ziqi Tan, Wulan Guo, Kai Zhao, Hongyun Yang, Mei Luo, Qi Ning, Feng Yang},pdftitle={Budget Boundary Effects in Test-Time Mathematical Reasoning}}
\title{Budget Boundary Effects in Test-Time\\
Mathematical Reasoning}
\author{%
  Guilin Zhang \quad Ziqi Tan \quad Wulan Guo \quad Kai Zhao \\
  \textbf{Hongyun Yang$^{1}$ \quad Mei Luo$^{1}$ \quad Qi Ning$^{1}$ \quad Feng Yang$^{1}$} \\[3pt]
  $^{1}$Yantong AI
}
\newcommand{\strict}{\mathrm{S}}
\newcommand{\advisory}{\mathrm{A}}
\newcommand{\maj}{\mathrm{Maj}}
\newcommand{\oracle}{\mathrm{Cov}}

\begin{document}
\maketitle
\begin{abstract}
A cumulative token cap can fall inside a mathematical derivation, forcing a test-time controller to choose between stopping at the cap (strict) and allowing the current attempt to finish (advisory). We measure this boundary choice with paired offline replays of 19,200 public traces: 120 AIME, BrUMO and HMMT problems and two archive configurations of one model. Candidate order and a 16-attempt cap are fixed, and answer selection is blind to reference answers and correctness labels. Three findings emerge. First, at the 4k cap, most advisory accuracy gains replace abstention with a correct answer; strict stopping pays for an unfinished prefix that the completed-only selector cannot use. Second, comparisons along realized cost differ from same-cap comparisons: advisory 4k in low has higher accuracy than strict 8k at comparable mean completion cost, while in high its observed accuracy is 0.42 points below strict 32k using 59\% of its mean tokens. These aggregate comparisons do not establish equal-compute superiority or accuracy equivalence. Third, increased candidate coverage does not guarantee higher answer accuracy: a log-probability selector loses accuracy while coverage rises, including after a source-grade consistency repair. Same-cap majority-accuracy differences shrink below 1.3 percentage points at 32k. Budget curves should jointly state the cap, realized cost, eligible candidates, stopping rule and selector information.
\end{abstract}

\section{Introduction}

A mathematical solver can spend its budget on intermediate reasoning \citep{wei2022cot,kojima2022zero}, sampled solutions \citep{wang2023consistency}, or tree search \citep{yao2023thoughts}. Compute allocation \citep{snell2025scaling,wu2025inference} and length control \citep{muennighoff2025s1,han2025token} motivate a practical accounting question: what happens when a cumulative cap falls inside the current attempt?

A budgeted mathematical agent must specify when its controller checks expenditure and which interrupted outputs remain usable. A strict controller stops generation at the cap; an advisory controller checks between attempts and allows the current one to finish. The same nominal budget can therefore yield different costs and candidate pools. Our sequential generate--select loop is a minimal setting for studying this decision. For search and verifier-based agents, recording step boundaries, costs and answer eligibility is a starting point for auditing budget enforcement.

Outcome verification \citep{cobbe2021verifiers}, process supervision \citep{lightman2024verify}, and generative verification \citep{singhi2025verify} target reasoning or selection. Our audit holds the archived attempts and selector fixed and measures how stopping changes answer availability and the returned decision. Short-budget AutoML audits \citep{zhang2026peeking} and calibrated resource-control comparisons \citep{zhang2026rlscale} motivate this focus on the connection between expenditure and evaluation.

\begin{figure}[t]
\centering
\includegraphics[width=\linewidth]{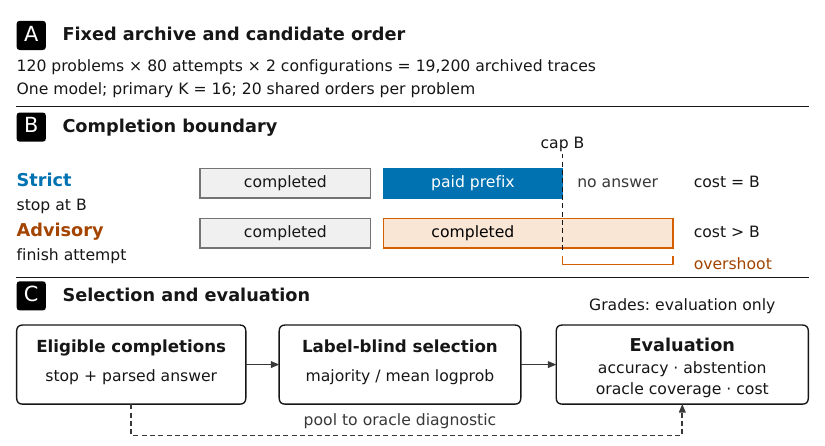}
\caption{Paired offline evaluation. A fixes the archived bank and order; B shows a boundary-crossing attempt under each completed-only rule; C applies the same eligibility and selection separately to each pool. Grades enter evaluation only; coverage is an oracle diagnostic. Exact boundaries or bank exhaustion need not overshoot. Timeline widths are illustrative, not to scale. Costs are completion tokens.}
\label{fig:framework}
\end{figure}

Building on the source study's distinction between completed-bank diagnostics and end-to-end inference \citep{hariri2026scaling}, Figure~\ref{fig:framework} organizes our audit: \textbf{(A)} pair each problem, archive configuration and order; \textbf{(B)} charge interrupted work and overshoot while varying the stopping rule; \textbf{(C)} separate candidate coverage, label-blind selection and evaluation. This reveals three distinct effects: abstention drives short-budget gains, realized cost changes cross-budget comparisons, and coverage can rise while selected-answer accuracy falls.

\section{A paired boundary audit}
\label{sec:protocol}

For one problem and one predetermined order, let $c_i>0$ be the recorded completion-token cost of attempt $i$, $p_i$ its prompt-token count, and $s_j=\sum_{i=1}^{j}c_i$, with $s_0=0$. We allow at most $K$ attempts. The replay reads complete archives to reconstruct costs, but its simulated controller launches an attempt using only the tokens spent so far; it never uses that attempt's eventual length, answer, or correctness to decide whether to launch it.

\paragraph{Stopping and eligibility.}
Under \emph{strict} stopping, generation ends when total completion cost reaches $B$. Attempt $i$ is available only if $s_i\leq B$. Under \emph{advisory} stopping, attempt $i$ is launched when $s_{i-1}<B$ and is allowed to finish. Thus, before parsing and finish-status filtering, the available index sets are
\begin{equation}
 I_{\strict}(B)=\{i\leq K:s_i\leq B\},\qquad
 I_{\advisory}(B)=\{i\leq K:s_{i-1}<B\}.
 \label{eq:pools}
\end{equation}
Only returned records with \texttt{finish\_reason=stop} and a valid parsed answer enter either selector. A strict boundary-crossing attempt contributes its consumed prefix tokens to the ledger and no answer to the pool. We intentionally study \emph{completed-only} reduction: we do not infer an answer from a truncated prefix or borrow the full archive's eventual answer. This is a declared terminal rule, not a claim that all prefixes lack usable answers.

\paragraph{Information available to selection.}
The primary selector takes the plurality of conservatively normalized boxed answers, conventionally termed majority vote, breaking ties by first occurrence and selecting the earliest representative of the winning answer. It reads only eligible generated text and order. References, archived correctness, and reference-assisted verifier scores are unavailable to it. The evaluator subsequently reads the selected representative's archived correctness label; an empty pool is an incorrect abstention, retained in the denominator. We extract the last boxed answer only if its braces balance and normalization is nonempty, preserving expressions such as $2^{99}$ and removing superficial spacing and wrappers. This does not merge all mathematically equivalent expressions. Archive extraction and a finite completion-mean-log-probability selector are sensitivity analyses.

We report \emph{candidate coverage}, $\oracle(I)=\mathbf{1}\{\exists i\in I:z_i=1\}$ for archived correctness labels $z_i$, as an explicitly reference-assisted diagnostic. Such any-correct metrics, including appropriately defined pass@$k$, legitimately describe a candidate pool; they do not report the performance of a reference-free final-answer selector \citep{brown2024monkeys}. Unlike fixed-pool selection diagnostics \citep{hu2026oracle}, our stopping intervention changes the eligible pool before reduction.

\paragraph{What the pairing identifies.}
The advisory pool adds at most one boundary attempt to the strict pool. Consequently coverage cannot decrease, while the majority decision can improve or worsen. We estimate the paired accuracy change $\Delta_{\maj}$, the coverage change $\Delta_{\oracle}$, and $\Delta_{\oracle}-\Delta_{\maj}$, the change in the coverage--selection gap. All compare the same problem and order. These quantities separate the availability of a correct answer from whether the declared selector returns it; they do not assume independent, additive ``budget'' and ``oracle'' effects.

\section{Data and experimental design}
\label{sec:data}

We reuse the public Scorio Trace archive \citep{hariri2026scaling}, with all 30 problems in each of AIME24, AIME25, BrUMO25, and HMMT25 February: 120 problems in total. The two archive configurations are \texttt{gpt-oss-20b\_low} and \texttt{gpt-oss-20b\_high}, each with 80 archived seeds (1234--1313) per problem, for 19,200 existing attempts. We generate no new model outputs. The rendered system prompt says \texttt{Reasoning: medium}, while developer prompts carry low/high instructions. We therefore treat low/high as archive labels, not verified causal interventions on an API effort parameter, and make policy comparisons within each configuration.

The primary analysis fixes $K=16$ and $B\in\{4,8,16,32\}\times 10^3$ completion tokens. Twenty predetermined random orders per problem sample the candidate bank without reading answers or labels; each policy sees the identical order. We average orders within each problem, then compute paired percentile bootstrap intervals by resampling problems within each competition (5,000 replicates), preserving the four equal-sized strata and configuration pairing. Orders are not additional independent problems. Canonical order and $K\in\{8,32\}$ are sensitivity analyses, with numerical summaries in Appendix~\ref{app:diagnostics}.

Strict realized completion cost is $\min(B,s_K)$, including a discarded partial attempt; advisory cost includes the final returned attempt. Prompt tokens are accounted for separately for every launch under a no-cache assumption. Majority voting adds no model-token calls. Invalid outputs and parser failures retain their cost and their problem's denominator.

\suppressfloats[t]
\section{Results and controls}
\label{sec:results}
\begin{table}[t]
\centering
\caption{Paired boundary audit: 120 problems, $K=16$, 20 orders. S: strict; A: advisory. Accuracy and oracle coverage use archived grades. The paired accuracy change is in percentage points, with a 95\% question-bootstrap interval. Cost is mean realized completion tokens divided by $B$; these are not matched-compute comparisons. Bank exhaustion can give cost/$B<1$.}
\label{tab:main}
\small
\setlength{\tabcolsep}{3pt}
\begin{tabular*}{\linewidth}{@{\extracolsep{\fill}}lrrrlrrrr@{}}
\toprule
 & & \multicolumn{2}{c}{Accuracy (\%)} & & \multicolumn{2}{c}{Coverage (\%)} & \multicolumn{2}{c}{Cost/$B$} \\
\cmidrule(lr){3-4}\cmidrule(lr){6-7}\cmidrule(l){8-9}
Archive & $B$ & S & A & $\Delta$ [95\% CI] & S & A & S & A \\
\midrule
low & 4k & 37.25 & 50.13 & 12.88 [9.79, 16.17] & 41.08 & 58.67 & 0.99 & 1.87 \\
 & 8k & 46.96 & 52.79 & 5.83 [4.08, 7.79] & 55.13 & 63.88 & 0.99 & 1.43 \\
 & 16k & 52.83 & 55.50 & 2.67 [1.75, 3.67] & 66.04 & 69.00 & 0.96 & 1.17 \\
 & 32k & 57.00 & 57.63 & 0.63 [0.21, 1.08] & 71.08 & 71.96 & 0.81 & 0.89 \\
\midrule
high & 4k & 18.54 & 61.58 & 43.04 [37.08, 49.17] & 18.54 & 62.25 & 1.00 & 4.68 \\
 & 8k & 33.71 & 61.92 & 28.21 [23.67, 33.04] & 34.13 & 63.00 & 1.00 & 2.66 \\
 & 16k & 47.21 & 62.42 & 15.21 [12.42, 18.25] & 47.92 & 63.75 & 1.00 & 1.65 \\
 & 32k & 62.00 & 63.25 & 1.25 [0.75, 1.83] & 63.21 & 64.75 & 1.00 & 1.19 \\
\bottomrule
\end{tabular*}
\end{table}

\paragraph{An answer-availability effect at short budgets.}
At $B=4$k, advisory stopping changes majority accuracy from 37.25\% to 50.13\% in low and from 18.54\% to 61.58\% in high (Table~\ref{tab:main}; full curves in Appendix Figure~\ref{fig:budget}). These same-cap differences are gains from additional computation, not equal-cost improvements. At 32k the changes shrink to 0.63 and 1.25 percentage points (pp). The low-archive 4k gain is concentrated in AIME24/25 (25.00/23.33 pp), versus BrUMO/HMMT (1.83/1.33 pp).

All 43.04 pp of improvement in high at 4k replaces abstention with a correct answer; no correct vote is harmed. Abstention falls from 80.54\% to 32.00\%. The remaining 32\% reflects first attempts hitting the archive's 32,768-token generation cap. Advisory returns exactly one attempt in 80.50\% of these high-4k replays, and its mean eligible pool has only 0.94 answers. Thus the largest effect primarily measures whether a first answer becomes available, rather than a gain from multi-attempt aggregation. For low, abstention-to-correct contributes 11.54 pp, correction of a wrong answer 1.50 pp, and harmed correct votes subtract 0.17 pp.

\paragraph{Realized cost changes the cross-budget comparison.}
In low, advisory 4k achieves 50.13\% at 7,485 mean tokens, exceeding strict 8k's 46.96\% at 7,940 tokens. The paired accuracy difference is 3.17 pp [1.42, 5.04], but the completion-cost difference interval includes zero. In high, advisory 4k achieves 61.58\% at 18,729 tokens: 14.38 pp above strict 16k at 16,000 tokens, and 0.42 pp below strict 32k's 62.00\% at 31,867 tokens, using 58.77\% of its mean cost. This last accuracy interval, $[-1.17,0.25]$ pp, does not establish equivalence. At 4k, interrupted prefixes account for 39.83\% (low) and 85.90\% (high) of strict token expenditure without contributing an eligible answer. Appendix~\ref{app:realized} reports cross-budget intervals and descriptive interpolation. These comparisons do not hold each problem's compute fixed; Appendix~\ref{app:diagnostics} adds prompt-inclusive costs and overshoot tails.

\paragraph{Extra candidates can hurt a reference-free selector.}
At 4k in low, coverage rises by 17.58 pp but majority accuracy rises by 12.88 pp, widening the coverage--selection gap by 4.71 pp. The gap change is 0.67 pp in high. At 32k in low, maximum mean log-probability loses 0.71 pp (95\% CI $[-1.29,-0.17]$), although coverage rises by 0.88 pp. It fixes 14 of 2,400 problem--order outcomes (0.58\%) and harms 31 (1.29\%). All 31 harms change the normalized answer, and all survive the evaluation-only grade repair described below. Increased availability of correct candidates therefore need not translate into higher returned-answer accuracy.

\paragraph{Controls and annotation limits.}
Identical completed-prefix and aligned synthetic-cost controls give identical predictions. Archive extraction gives the same 4k majority change for low and 43.00 pp for high. Eight normalized-answer groups have conflicting source grades (two low, six high; all HMMT). An evaluation-only repair against normalized accepted reference forms resolves these conflicts and changes paired majority effects by at most 0.08 pp (low) and 0.33 pp (high); selectors remain unchanged. Appendix~\ref{app:repro} details these checks.

\section{Scope and implications}

Completed answers are discrete events inside a token ledger. A low score may reflect no complete derivation, wrong answers, or selection failure. Completing an attempt addresses availability; improving the reducer addresses selection. Length control with reinforcement learning \citep{aggarwal2025l1}, dynamic early exit \citep{yang2026exit}, and answering without extended thinking \citep{ma2025without} offer alternatives that change generation or the terminal rule and require their own trajectories.

Adaptive early stopping \citep{sun2025stop} detects redundant reasoning; optimized sample allocation \citep{zhang2025allocation} redistributes compute; stochastic backtracking \citep{tran2026backtracking} revisits stored prefixes. Our fixed-order replay isolates an accounting choice these controllers must also specify. It does not compare their policies. Reconstructed prefix boxes (Appendix~\ref{app:diagnostics}) further motivate evaluating prefix-aware terminal rules.

The estimates concern one model's two archives and a completed-only, sequential controller with shared order and eligibility. Normalization can split equivalent expressions; consistency checks only partly address source-grade errors. The results neither rank adaptive agents nor establish an online matched-compute advantage. Following the source study's full-system reporting requirements \citep{hariri2026scaling}, budget curves should state the cap, realized cost, attempted and eligible counts, stopping rule and selector information jointly.

\clearpage
\bibliographystyle{plainnat}
\begingroup
\small
\raggedright
\interlinepenalty=10000
\setlength{\bibsep}{1pt}
\bibliography{references}
\endgroup

\clearpage
\appendix
\section{Boundary identities and reproducibility}
\label{app:repro}

\begin{figure}[!ht]
\centering
\includegraphics[width=\linewidth]{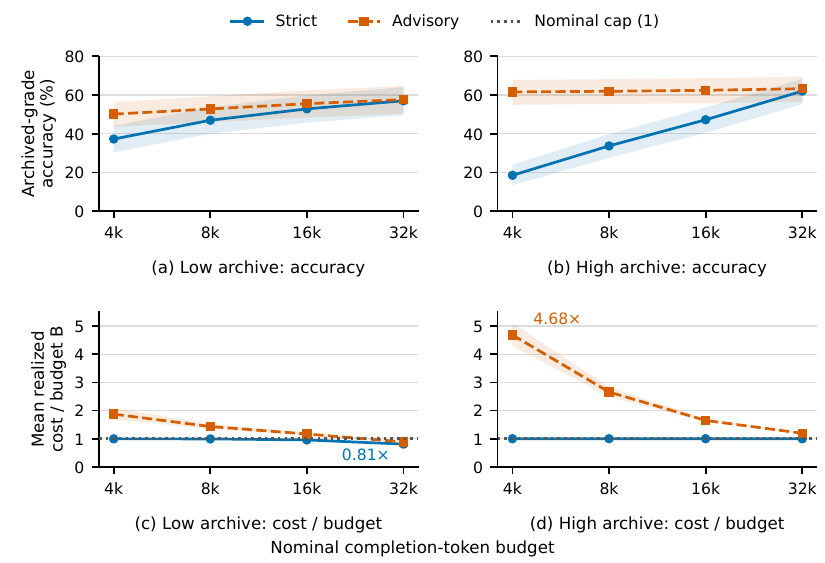}
\caption{Strict and advisory majority accuracy (top) and realized completion cost divided by the nominal cap $B$ (bottom), with $K=16$. Curves average 20 orders within each of 120 problems; shading gives problem-bootstrap 95\% intervals for each curve, not the paired difference. Bottom-row intervals are also divided by $B$. Dotted lines at one mark the nominal cap; bank exhaustion can yield ratios below one.}
\label{fig:budget}
\end{figure}

\paragraph{One extra attempt.}
If $s_{j-1}<B<s_j$ for some $j\leq K$, strict stopping returns the preceding $j-1$ attempts; advisory stopping also returns attempt $j$, and then launches none. If the budget is reached exactly at a completion boundary, or the bank is exhausted before the cap, both policies return the same attempts. Applying the same finish-status and answer parser to the two sets preserves their inclusion relation and a difference of at most one eligible answer. This proves the coverage monotonicity used in the audit; it imposes no monotonicity on majority accuracy.

\paragraph{Paired events.}
For selected-answer correctness $M_{\strict},M_{\advisory}\in\{0,1\}$, the average accuracy difference is
\[
 \Delta_{\maj}=\Pr(M_{\strict}=0,M_{\advisory}=1)
                   -\Pr(M_{\strict}=1,M_{\advisory}=0).
\]
Both events are meaningful: one additional answer can rescue a failed vote or overturn a correct one. In contrast, $\Delta_{\oracle}=\Pr(\oracle_{\strict}=0,\oracle_{\advisory}=1)$ because the coverage loss event is impossible for nested pools.

\paragraph{Reproduction record.}
The audit uses public traces from Scorio Trace \citep{hariri2026scaling}. This appendix records the pinned revision, retained questions and seeds, parsing and finish-status rules, controls, and bootstrap settings. Generated-answer fields passed to selection are separate from reference fields used by evaluation. Downloading and replaying archived inputs does not generate new reasoning attempts. The data card declares an MIT license for the trace release while retaining the original terms for competition problem statements.

\paragraph{Pinned inputs.}
Revision \texttt{176c59089b2367d569e1062ffe94a890bf3cbd17} is pinned. For each configuration and competition, the source shard contains 2,400 records, comprising 30 questions and 80 seeds. The primary parser locates the last \verb|\boxed{| occurrence, balances braces from that occurrence, and rejects a missing or unfinished box. Normalization removes whitespace, math delimiters and \verb|\left|/\verb|\right|, maps \verb|\dfrac| and \verb|\tfrac| to \verb|\frac|, unwraps whole-expression text/style wrappers, and canonicalizes literal integers. It does not simplify algebraic expressions, compare against a reference, or propagate correctness between candidates.

\begin{table}[h]
\centering
\caption{Source-record audit, each row containing 2,400 attempts. ``No box'' includes records without a complete nonempty parsed final box; ``Length'' records remain costly but ineligible, even if a box appears earlier. The columns overlap and must not be added as disjoint exclusions. Agreement compares primary boxed normalization with archive extraction, with no correctness-based filtering.}
\label{tab:audit}
\small
\begin{tabular*}{\linewidth}{@{\extracolsep{\fill}}llrrrr@{}}
\toprule
Archive & Competition & No box & Length & Parser agreement & Median tokens \\
\midrule
low & AIME24 & 132 & 122 & 2268 & 2924.0 \\
low & AIME25 & 123 & 121 & 2277 & 3471.0 \\
low & BrUMO25 & 16 & 0 & 2307 & 1180.0 \\
low & HMMT25 & 34 & 1 & 2366 & 1307.5 \\
\midrule
high & AIME24 & 400 & 413 & 2000 & 6994.0 \\
high & AIME25 & 508 & 543 & 1892 & 8880.0 \\
high & BrUMO25 & 736 & 841 & 1584 & 19183.0 \\
high & HMMT25 & 1152 & 1333 & 1248 & 32768.0 \\
\bottomrule
\end{tabular*}
\end{table}

\paragraph{A truncation floor in the high archive.}
At every tested budget, high-archive advisory abstention is 32.00\%. The same 768 of 2,400 problem--order pairs begin with a length-capped response costing 32,768 tokens, which is greater than every tested $B$. It remains ineligible and prevents a second attempt under advisory stopping. This is an observed consequence of the inherited generation cap and our eligibility rule, not a parsing bug or an estimate of the model's irreducible failure rate.

\paragraph{Source-grade sensitivity.}
Conflicts occur when the same conservatively normalized answer receives both correct and incorrect archive labels within a problem. We find two such groups in low HMMT and six in high HMMT. An exploratory evaluation-only repair sets a label to correct when it was already source-correct \emph{or} the generated normalized box exactly equals one of the normalized accepted reference forms. This adjusts 17 low and 41 high candidate labels and resolves all eight detected groups. It does not claim complete symbolic grading and cannot repair false-positive source labels. Crucially, repaired labels are read only after selection and never change the candidate pool, its order, or the answer returned. Table~\ref{tab:grading} retains the primary source-grade estimate beside this sensitivity analysis.

\begin{table}[h]
\centering
\caption{Paired majority change (advisory minus strict, pp) under primary source grades and exploratory formatting-repaired grades. The repaired values are a robustness check, not replacements for the prespecified source-grade table.}
\label{tab:grading}
\small
\begin{tabular*}{\linewidth}{@{\extracolsep{\fill}}lrrrr@{}}
\toprule
Archive & $B$ & Source grade & Repaired grade & Change in effect \\
\midrule
low & 4k & 12.875 & 12.958 & 0.083 \\
 & 8k & 5.833 & 5.917 & 0.083 \\
 & 16k & 2.667 & 2.667 & 0.000 \\
 & 32k & 0.625 & 0.625 & 0.000 \\
\midrule
high & 4k & 43.042 & 43.375 & 0.333 \\
 & 8k & 28.208 & 28.458 & 0.250 \\
 & 16k & 15.208 & 15.333 & 0.125 \\
 & 32k & 1.250 & 1.292 & 0.042 \\
\bottomrule
\end{tabular*}
\end{table}

\begin{figure}[t]
\centering
\includegraphics[width=\linewidth]{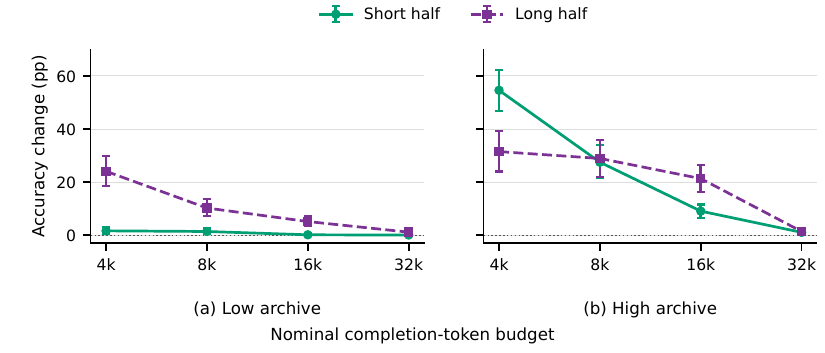}
\caption{Paired majority changes (advisory minus strict, percentage points) in each archive's shorter and longer halves. Points average the 20 paired orders within each problem; intervals use the same stratified question-bootstrap procedure. Length groups differ in task and difficulty, so these are not causal effects of making a solution longer.}
\label{fig:length}
\end{figure}

\paragraph{Controls and length strata.}
Exact completed-prefix controls and aligned equal-cost controls yield zero prediction disagreements across 60,480 paired selector comparisons, including canonical order. These controls verify the declared coupling; they do not establish an advantage at matched realized compute. For length strata, we rank the 120 questions within each archive by the median cost of all 80 candidates, break ties deterministically, and split into halves of 60 without reading correctness. Competition and difficulty composition may differ across the halves. Their curves in Figure~\ref{fig:length} are therefore descriptive conditional results.

\begin{samepage}
At 4k in high, the longer half gains 31.50 pp and spends an extra 23,697 completion tokens on average, whereas the shorter half gains 54.58 pp with an extra 5,761 tokens. Longer archived traces are therefore not uniformly favored: many hit the original generation limit and remain ineligible. In low, the corresponding gains are 24.08 pp for the longer half and 1.67 pp for the shorter half.
\par
\end{samepage}

\section{Exploratory comparisons along realized cost}
\label{app:realized}

These post-hoc summaries use the frozen primary replay outputs: $K=16$, boxed-answer majority, archived grades, all 120 problems and orders 1--20. They add no model calls or replay budgets. We average orders within each problem and use 5,000 paired bootstrap resamples stratified by competition, with seed 20260905. The same resample determines costs and accuracy in every condition. Intervals are pointwise and unadjusted for multiplicity.

\begin{table}[h]
\centering
\caption{Existing budget points compared by mean realized completion cost. A always uses a nominal 4k cap; the S cap varies. Accuracy and the paired A minus S difference use archived grades. These are different-cost comparisons.}
\label{tab:costcomparisons}
\small
\setlength{\tabcolsep}{3pt}
\begin{tabular*}{\linewidth}{@{\extracolsep{\fill}}llrrrrl@{}}
\toprule
 & & \multicolumn{2}{c}{Mean tokens} & \multicolumn{2}{c}{Accuracy (\%)} & \\
\cmidrule(lr){3-4}\cmidrule(lr){5-6}
Archive & S cap & A & S & A & S & $\Delta$ pp [95\% CI] \\
\midrule
low & 8k & 7,485 & 7,940 & 50.13 & 46.96 & 3.17 [1.42, 5.04] \\
high & 16k & 18,729 & 16,000 & 61.58 & 47.21 & 14.38 [11.54, 17.50] \\
 & 32k & 18,729 & 31,867 & 61.58 & 62.00 & $-0.42$ [$-1.17$, 0.25] \\
\bottomrule
\end{tabular*}
\end{table}

The observed low cost difference is $-454$ tokens (95\% CI $[-1{,}202,384]$), so its lower sample mean does not establish a population cost reduction. In high, the advisory-4k/strict-32k mean-cost ratio is 0.588 [0.539, 0.640]; the accuracy interval containing zero does not establish equivalence. Mean-cost comparisons leave the allocation across problems uncontrolled.

\paragraph{Interpolation of aggregate means.}
Let $(x_b,y_b)$ denote the strict mean-token cost and accuracy at nominal budget $b$. At advisory 4k's mean cost $x_A$, we select adjacent strict points by cost and compute $\widetilde y_S=(1-w)y_L+w y_R$, where $w=(x_A-x_L)/(x_R-x_L)$. The strict endpoints are 4k--8k in low and 16k--32k in high. Interpolated accuracies are 45.85\% and 49.75\%, giving advisory-minus-interpolated differences of 4.28 pp [2.13, 6.71] and 11.83 pp [9.27, 14.62]. Each bootstrap resample recomputes all means, cost-selected neighbors and weights; none requires extrapolation. Low uses 8k--16k neighbors in 703 of 5,000 resamples. These intervals exclude uncertainty in the assumed linear curve shape. Only the endpoints of each segment are observed: a concave strict curve would lie above its chord and reduce the advisory gap; convexity reverses this direction. The lines connecting nominal-budget points in Figure~\ref{fig:budget} do not establish local linearity on the realized-cost axis. Interpolation does not evaluate a controller at matched per-problem compute.

\begin{table}[h]
\centering
\caption{Strict cost assigned to unfinished terminal prefixes. $U$ is mean charged prefix tokens; $f$ divides total prefix tokens by total strict cost over 2,400 problem--order replays. It excludes fully returned but ineligible responses and is not a mean of per-replay fractions.}
\label{tab:prefixcost}
\small
\begin{tabular*}{\linewidth}{@{\extracolsep{\fill}}lrrrr@{}}
\toprule
 & \multicolumn{2}{c}{low} & \multicolumn{2}{c}{high} \\
\cmidrule(lr){2-3}\cmidrule(l){4-5}
$B$ & $U$ (tokens) & $f$ (\%) & $U$ (tokens) & $f$ (\%) \\
\midrule
4k & 1,585 & 39.83 & 3,436 & 85.90 \\
8k & 2,138 & 26.92 & 5,873 & 73.42 \\
16k & 2,875 & 18.80 & 9,879 & 61.74 \\
32k & 3,436 & 13.29 & 15,095 & 47.37 \\
\bottomrule
\end{tabular*}
\end{table}

The prefixes in Table~\ref{tab:prefixcost} cannot contribute candidates under this completed-only rule. This does not show that their mathematical content is unusable by a different terminal rule. Advisory stopping can still return a length-capped or invalid response and incur cost without an eligible answer. Appendix~\ref{app:diagnostics} examines answer availability inside reconstructed prefixes separately from these completed-only estimates.

\section{Additional accounting and selection diagnostics}
\label{app:diagnostics}

These exploratory additions were computed after workshop acceptance from the same archived traces and frozen replay outputs. They introduce no new generation, controller tuning or primary test set. Each configuration contains 2,400 primary problem--order replays per budget; orders remain repeated measurements of 120 problems.

\paragraph{Effective pool size and overshoot tails.}
Table~\ref{tab:counts} reports eligible pools after finish-status and parsing filters and the empirical 95th percentile of advisory completion cost. The percentiles describe the pooled replay distribution, not confidence bounds or guaranteed caps. At high 4k, advisory returns exactly one attempt in 80.50\% of replays; its eligible-pool distribution is 768 empty, 1,174 single-answer and 458 multi-answer pools. Low 4k has 54 empty, 480 single-answer and 1,866 multi-answer pools. A nominal 4k cap therefore describes very different effective sampling regimes across the two archives. Tail costs also substantially exceed mean costs.

\begin{table}[!ht]
\centering\small
\caption{Effective pools and advisory tail cost, $K=16$. S and A denote strict and advisory. ``One return'' includes a single returned attempt even if it is length-capped or has no valid box. Eligible counts exclude such responses. P95 is the 95th percentile across 2,400 replay costs.}
\label{tab:counts}
\begin{tabular*}{\linewidth}{@{\extracolsep{\fill}}llrrrrr@{}}
\toprule
Archive & $B$ & S eligible & A eligible & A one return (\%) & A P95 tokens & P95/$B$ \\
\midrule
low & 4k & 2.05 & 3.01 & 21.21 & 25,042 & 6.26 \\
low & 8k & 4.28 & 5.21 & 11.17 & 29,654 & 3.71 \\
low & 16k & 7.88 & 8.68 & 6.04 & 32,768 & 2.05 \\
low & 32k & 11.44 & 11.86 & 2.29 & 46,855 & 1.46 \\
high & 4k & 0.26 & 0.94 & 80.50 & 32,768 & 8.19 \\
high & 8k & 0.68 & 1.34 & 64.29 & 32,768 & 4.10 \\
high & 16k & 1.52 & 2.15 & 49.50 & 37,804 & 2.36 \\
high & 32k & 3.13 & 3.65 & 32.71 & 58,348 & 1.82 \\
\bottomrule\end{tabular*}\end{table}

\paragraph{Prompt-inclusive accounting.}
At the same nominal budget and order, both rules launch exactly the same attempts: strict pays for the final prompt before interrupting that attempt, while advisory allows it to finish. We verify identical prompt counts for every primary pair. Under the no-cache token ledger, adding prompts therefore leaves every same-cap cost difference unchanged. Cross-budget comparisons do launch different numbers of attempts. Including prompts, low advisory 4k averages 8,100 total tokens versus 8,974 for strict 8k (paired difference $-874$, 95\% CI $[-1{,}646,-22]$). High advisory 4k averages 18,993 versus 16,510 for strict 16k and 32,695 for strict 32k. These are token totals; input and output pricing, caching, serving latency and FLOPs are not modeled.

\paragraph{Boxes already present in interrupted prefixes.}
The completed-only rule excludes every interrupted attempt, including one that has already emitted a box. We audit this distinction using the model's official tokenizer.\footnote{Tokenizer revision: \texttt{6cee5e81ee83917806bbde320786a8fb61efebee}, from \url{https://huggingface.co/openai/gpt-oss-20b}.} The archive provides rendered text and recorded completion-token counts, rather than the original generated token-ID stream. We retokenize the complete text and examine a prefix at the strict ledger's remaining token count. We retain only traces whose recorded count equals the retokenized count or exceeds it by one, and treat other differences as unreconciled. This count check does not prove positional alignment; the analysis is a retokenized-text sensitivity, not a reconstruction of an observed online cancellation.

Table~\ref{tab:prefixboxes} separates examined from unverified boundary replays. ``Any box'' means at least one balanced nonempty box anywhere in the reconstructed prefix; ``Last box'' applies the primary parser, which rejects an unfinished last box even when an earlier complete box exists. At 4k, 1.78\% of examined low prefixes and 6.43\% of examined high prefixes contain any complete box. Thus usable-looking prefixes do exist. We do not transfer a full attempt's eventual correctness label to an earlier box or estimate prefix-aware accuracy from these counts. Unverified traces may have different box frequencies.

\begin{table}[!ht]
\centering\small
\caption{Answer-format availability in reconstructed strict boundary prefixes. Counts are replay events and may reuse a trace across orders. Unverified = boundary events minus examined events. No correctness claim is made for prefix boxes.}
\label{tab:prefixboxes}
\begin{tabular*}{\linewidth}{@{\extracolsep{\fill}}llrrrrr@{}}
\toprule
Archive & $B$ & Boundary events & Examined & Unverified & Any box & Last box \\
\midrule
low & 4k & 2,380 & 2,297 & 83 & 41 & 40 \\
low & 8k & 2,347 & 2,252 & 95 & 45 & 45 \\
low & 16k & 2,054 & 1,967 & 87 & 37 & 37 \\
low & 32k & 1,185 & 1,124 & 61 & 24 & 24 \\
high & 4k & 2,399 & 2,271 & 128 & 146 & 142 \\
high & 8k & 2,400 & 2,264 & 136 & 249 & 248 \\
high & 16k & 2,399 & 2,258 & 141 & 296 & 291 \\
high & 32k & 2,336 & 2,201 & 135 & 305 & 300 \\
\bottomrule\end{tabular*}\end{table}

\paragraph{Selection harm after grade repair.}
For low at 32k, maximum completion-mean log-probability fixes 14 outcomes and harms 31 of 2,400 under both source and repaired grades. All 31 harmful events change the normalized answer; none is solely a change in the selected representative of an identical normalized answer. The paired mean and interval remain $-0.71$ pp $[-1.29,-0.17]$. For example, AIME24 archive question ID 3, order 5, changes from seed 1279's answer 809 to seed 1271's answer 808; source and repaired evaluation both mark the former correct and the latter incorrect. For high at 32k, repair changes the counts from 41 fixes/16 harms to 40 fixes/14 harms, removing two same-answer harms and one same-answer fix. This illustrates why the representative-label audit matters even when the low-archive failure survives it.

\clearpage
\paragraph{Attempt cap and order sensitivity.}
Table~\ref{tab:ksensitivity} reports paired majority changes for the nested $K=8,16,32$ banks and ascending-seed canonical order. The high configuration seldom reaches a large attempt cap at these budgets, so changing $K$ has little effect. The ranges across the 20 orders quantify ordering sensitivity within this bank, not sampling uncertainty over problems. They must not be interpreted as confidence intervals.

\begin{table}[!ht]
\centering\small
\caption{Advisory minus strict majority accuracy (pp). The first three numeric columns average the same 20 predetermined orders per problem. Canonical uses ascending archive seed and $K=16$. The final column spans the 20 individual-order mean effects at $K=16$.}
\label{tab:ksensitivity}
\begin{tabular*}{\linewidth}{@{\extracolsep{\fill}}llrrrrl@{}}
\toprule
Archive & $B$ & $K=8$ & $K=16$ & $K=32$ & Canonical & Order range \\
\midrule
low & 4k & 12.875 & 12.875 & 12.875 & 15.000 & [5.833, 17.500] \\
low & 8k & 5.792 & 5.833 & 5.833 & 8.333 & [0.000, 10.000] \\
low & 16k & 2.708 & 2.667 & 2.667 & 5.833 & [0.000, 5.833] \\
low & 32k & 0.542 & 0.625 & 0.667 & 1.667 & [-0.833, 1.667] \\
high & 4k & 43.042 & 43.042 & 43.042 & 37.500 & [39.167, 48.333] \\
high & 8k & 28.208 & 28.208 & 28.208 & 23.333 & [22.500, 34.167] \\
high & 16k & 15.208 & 15.208 & 15.208 & 14.167 & [11.667, 20.833] \\
high & 32k & 1.250 & 1.250 & 1.250 & 2.500 & [0.000, 4.167] \\
\bottomrule\end{tabular*}\end{table}

\paragraph{Order and resampling details.}
Each bank is first sorted by archive seed. For competition index $t=0,1,2,3$ in the order AIME24, AIME25, BrUMO25, HMMT25 February, archive question ID $q=0,\ldots,29$, and repeat $r=1,\ldots,20$, NumPy's default PCG64 generator is initialized through \texttt{SeedSequence([20260904, t, q, r])}. It permutes all 80 positions, of which the first $K$ form the nested bank. The same permutation is used in both configurations. Bootstrap seed 20260905 generates 5,000 multinomial count vectors within each 30-question competition, with the same counts applied to paired conditions. The full-set design followed an exploratory 20-question low-archive feasibility pilot; the post-acceptance diagnostics above remain exploratory.

\end{document}